\documentclass[9pt]{Interspeech}
\interspeechcameraready

\title{End-to-End Speech Translation for Low-Resource Languages Using Weakly Labeled Data}
\author[affiliation={1}]{Aishwarya}{Pothula}
\author[affiliation={1}]{Bhavana}{Akkiraju}
\author[affiliation={1}]{Srihari}{Bandarupalli}
\author[affiliation={1}]{Charan}{D}
\author[affiliation={2}]{Santosh}{Kesiraju}
\author[affiliation={1}]{Anil Kumar}{Vuppala}
\affiliation{Speech Processing Laboratory}{International Institute of Information Technology, Hyderabad}{India}
\affiliation{Speech@FIT}{Brno University of Technology}{Czechia}
\email{\{aishwarya.pothula, bhavana.akkiraju\}@research.iiit.ac.in}

\usepackage{comment}
\usepackage{cite}
\usepackage{comment}
\usepackage{amsmath,amssymb,amsfonts}
\usepackage{algorithmic}
\usepackage{graphicx}
\usepackage{textcomp}
\usepackage{booktabs}
\usepackage{xcolor}
\usepackage{hyperref}
\usepackage{tikz}
\usetikzlibrary{bayesnet}
\usetikzlibrary{arrows}
\def\BibTeX{{\rm B\kern-.05em{\sc i\kern-.025em b}\kern-.08em
    T\kern-.1667em\lower.7ex\hbox{E}\kern-.125emX}}
\usepackage{graphicx}
\begin{document}
%\bstctlcite{IEEEexample:BSTcontrol}

\maketitle
\begin{abstract}
    
    The scarcity of high-quality annotated data presents a significant challenge in developing effective end-to-end speech-to-text translation (ST) systems, particularly for low-resource languages. This paper explores the hypothesis that weakly labeled data can be used to build ST models for low-resource language pairs. We constructed speech-to-text translation datasets with the help of bitext mining using state-of-the-art sentence encoders. We mined the multilingual Shrutilipi corpus to build Shrutilipi-anuvaad, a dataset comprising ST data for language pairs Bengali-Hindi, Malayalam-Hindi, Odia-Hindi, and Telugu-Hindi. We created multiple versions of training data with varying degrees of quality and quantity to investigate the effect of quality versus quantity of weakly labeled data on ST model performance. Results demonstrate that ST systems can be built using weakly labeled data, with performance comparable to massive multi-modal multilingual baselines such as SONAR and SeamlessM4T. 
\end{abstract}
\section{Introduction}

Recent years have seen an increased interest towards building speech translation (ST) systems for low-resource languages~\cite{iwslt:2023,iwslt-2024-findings}, enabled by transfer learning~\cite{StoianBG20,kesiraju_strategies_2023} from large pre-trained models~\cite{seamless2023,inaguma-etal-2023-unity}, and the availability of (smaller) training datasets across several languages. The widely used datasets for ST research are CoVoST~\cite{wang21s_interspeech} and MUST-C~\cite{di-gangi-etal-2019-must}, which are built on top of Mozilla common voice~\cite{ardila-etal-2020-common} and TED-talks respectively.  The aforementioned datasets mainly span English$\leftrightarrow$other languages, and relatively few datasets exist that are beyond English (e.\thinspace g: Tamasheq$\rightarrow$French~\cite{zanon-boito-etal-2022-speech}, Quechua$\rightarrow$Spanish~\cite{cardenas2018siminchik}). SpeechMatrix~\cite{duquenne-etal-2023-speechmatrix}, a large-scale multilingual dataset covering 136 language pairs, was created using recordings from European parliaments and expands the diversity of ST datasets by including non-English-centric translation pairs. The International Conference on Spoken Language Translation (IWSLT) is one of the long-standing, notable research communities that has been organizing an annual conference, along with shared tasks (scientific challenges) that encourage researchers and practitioners to participate in advancing state-of-the-art on speech translation technologies, spanning several domains and languages. Although some tracks release training and evaluation data with unrestricted usage of scientific research, others provide only a time-bound license (typically four months) for the data. To the best of our knowledge, there exist no freely available Indic-to-Indic speech translation datasets spanning multiple languages.

Developing speech translation systems for low-resource languages is challenging due to the lack of large and high-quality training datasets~\cite{jia22b_interspeech}. Our research seeks to bridge this gap by building \textit{ end-to-end speech-to-text translation models} using \textit{ weakly labeled data} that are automatically curated from multilingual speech datasets. For this purpose, we use Shrutilipi (SL)~\cite{Shrutilipi_2023}, a corpus of multilingual speech data for 12 Indian languages. SL contains speech samples from broadcast news along with (nearly accurate) text transcripts. Using the state-of-the-art multilingual sentence encoder model (SONAR~\cite{sonar:2023}), we find textual sentence pairs that are closer in the shared embedding space~\cite{Duquenne2021MultimodalAM}. Pairs with high similarity score are kept for development and test sets, whereas pairs with mid-to-low similarity scores are used for training ST systems; thereby creating a weakly labeled training dataset. Sentence encoders for mining bitexts has been extensively studied and applied in the machine translation community \cite{schwenk-etal-2021-ccmatrix, sloto-etal-2023-findings, chaudhary-etal-2019-low}.

Motivation for our work comes from three sources, \\
(i) previous work~\cite{wang21s_interspeech,sethiya-etal-2024-indic} that relied on sentence encoders for creating CoVoST and Indic-TEDST datasets, (ii) lack of Indic-to-Indic speech translation datasets, and (iii) the research question of building ST systems from weakly labeled automatically curated data. For the last point, we employ the transfer learning approach, where our ST models are initialized from pre-trained bilingual ASR models~\cite{kesiraju_strategies_2023}. We also use state-of-the-art multilingual massive speech-to-text models such as SONAR~\cite{sonar:2023} and Seamless~\cite{seamless2023} as baseline machine and speech translation (MT, ST) systems.

To summarize:
\begin{enumerate}
    \item We have curated an Indic to Indic speech translation dataset, dubbed \textit{Shrutilipi-anuvaad}, that offers itself as a crucial resource for spoken translation tasks for four language pairs: Bengali-Hindi (bn-hi), Malayalam-Hindi (ml-hi), Odia-Hindi (or-hi), and Telugu-Hindi (te-hi). Details are discussed in Section~\ref{sec:data}.
    \item Through our experiments (Sections~\ref{sec:exp} and \ref{sec:res}), we demonstrate that \textit{large amounts of weakly labeled data} can be leveraged by pre-trained ASR models, that result in ST models which perform comparably and often outperform the out-of-the-box SOTA (MT, ST) models like SONAR and Seamless. %, a significant finding in low-resource language scenarios where high-quality annotations are often limited.
    \item Further analysis (Section~\ref{sec:res}) provides valuable insights into the \textit{ quality and quantity} of weakly labeled data and its impact on ST performance; thus helping us to understand the trade-offs involved when using available datasets. The data and code are available at: \texttt{\url{https://github.com/aishwaryapothula/Shrutilipi-Anuvaad}}.
%     \item \textbf{Expanding Dataset Utility}:
% Our approach to curating the Shruthilipi dataset underscores the potential for expanding the utility of existing datasets for a broader range of speech-related applications. This addresses key challenges in processing low-resource languages.
\end{enumerate}

\section{Curating Shrutilipi-anuvaad dataset} 
\label{sec:data}
\begin{figure}[!ht]
    \centering
    \includegraphics[width=0.99\linewidth]{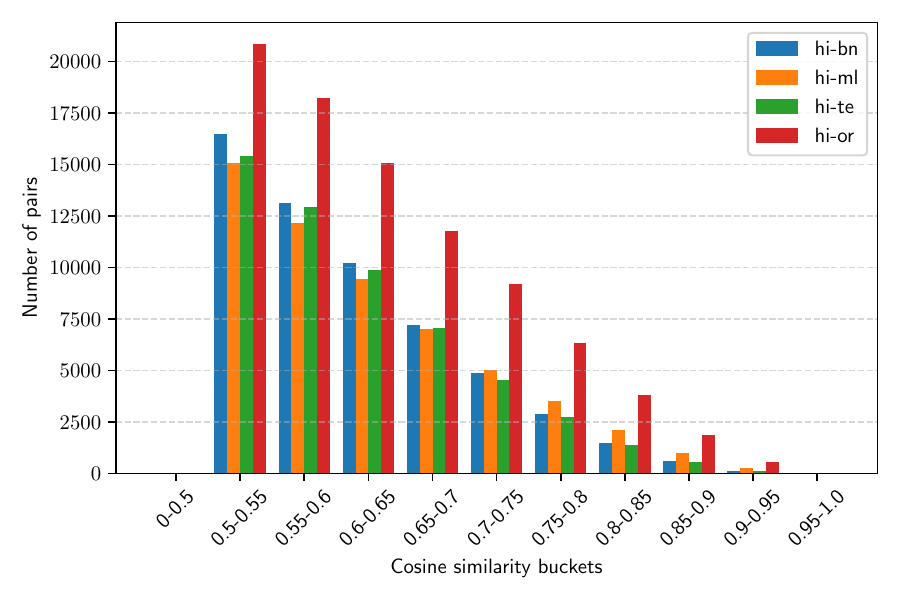}
    \caption{Histogram of cosine similarity scores computed on sentence embeddings extracted using SONAR.}
    \label{fig:hist}
\end{figure}
Shrutilipi is a multilingual speech dataset derived by mining audio and text pairs from All India Radio news broadcasts~\cite{Shrutilipi_2023}. We expected the presence of parallel data, since some of the news broadcasts describe the same events in multiple languages. Using SONAR text encoder, we extracted sentence embeddings for all transcripts and grouped text pairs based on cosine similarity. Pairs with a high similarity score ($>0.8$) were split equally into development and test sets; whereas pairs with mid-to-high ($0.5, 0.8$) similarity scores were used as weakly labeled training set. 15\% of the test data has been validated by native speakers of the languages and found to be largely accurate, with the majority (more than 80\%) receiving a score of 4 or 5 on a scale of 5, indicating high accuracy and alignment with human judgment. The resulting dataset contains quadruples (text pair and corresponding speech pair) for each language pair. 
\begin{table}[!ht]
    \caption{Statistics of the Shrutilipi-anuvaad dataset}
    \label{tab:slt_data_stats}
    \centering
    \footnotesize
    \resizebox{\columnwidth}{!}{%
    \begin{tabular}{crrrr} \toprule
 Lang. & \multicolumn{4}{c}{\# Utterances (hours)}\\
     pair  & Training $\mathcal{S}_1$ & Training $\mathcal{S}_5$ & Dev. & Test \\ \midrule
      hi-bn&  52k (78)& 8.8k (12)& 1.7k (2.4)& 1.7k (2.1)\\
 hi-ml& 52k (71)& 7.2k (9.4)&1.0k (1.3)&1.0k (1.3)\\
 hi-te& 54k (90)& 7.7k (11.6)&1.0k (1.8)&1.0k (1.6)\\
 hi-or& 81k (120)& 15k (24)& 3.1k (4.7)&3.1k (4.6)\\ \bottomrule
    \end{tabular}
    }
    %\vspace{-0.2cm}
\end{table}
\begin{table}[!ht]
    \label{tab:asr_data_st}
    \centering
    \caption{Statistics of the data for ASR training}
    \footnotesize
    \begin{tabular}{cccccc} \toprule
  Language &  hi & bn & ml & te & or \\ \midrule 
  \# Utterances & 662k & 176k & 185k & 173k & 225k \\ 
  Hours  & 934 & 248 & 242 & 311 & 326\\ \bottomrule
    \end{tabular}
    %\vspace{-0.2cm}
\end{table}

The histogram of similarity scores for four language pairs is shown in Fig.~\ref{fig:hist}. We can observe that most pairs are concentrated in the similarity bin$(0.5, 0.6)$. Pairs with score less than 0.5 are not considered to be part of ST training set, instead they were used for pre-training ASR systems. We ensured that there is no overlap of sentences between the ASR pre-training and ST dev and test sets, preventing data contamination.

Based on the similarity scores  in the range $(0.5, 0.8)$ the training set is further divided into five sets $\mathcal{S}_1 \supset \mathcal{S}_2 \supset \mathcal{S}_3 \supset  \mathcal{S}_4 \supset \mathcal{S}_5$. The largest set $\mathcal{S}_1$ contains all the pairs in the $(0.5, 0.8)$ score bin, whereas the smallest set $\mathcal{S}_5$ contains pairs from $(0.7, 0.8)$ bin. The other splits $\mathcal{S}_2, \mathcal{S}_3, \mathcal{S}_4$ are made from bins $(0.6, 0.8), (0.62, 0.8), (0.68, 0.8)$ respectively. 
The statistics of the resulting dataset is presented in Table~\ref{tab:slt_data_stats}. Although not given in the Table, $\mathcal{S}_2$ is almost half the size of $\mathcal{S}_1$ and $\mathcal{S}_5$ is about six times smaller than $\mathcal{S}_1$. The average utterance lengths across all splits $\mathcal{S}_1$-$\mathcal{S}_{5}$ and the dev/test sets for all language pairs are comparable (e.g., Hi–Te: $\mathcal{S}_1$-$\mathcal{S}_{5}$  = 11.4–11.8, Dev/Test = 11.7/10.9; Hi–Bn: 12.2–13.4, Dev/Test = 13.1/11.5), suggesting that there is no significant length-based bias in the bucketing process. The resulting dataset is dubbed \textit{Shrutilipi anuvaad}.

The final training data for the bilingual ASRs is created by combining sentence pairs with similarity scores between 0.5 and 0.8, along with their corresponding speech data. The left-over monolingual speech-text pairs (scores $< 0.5$) are also included. 
% The test and development data are generated from the same 0.8-1.0 split by mapping text of both source and target languages with corresponding speech data.
Although Shrutilipi contains 12 languages, our experiments and analysis in this paper are restricted to 4 language pairs: bn-hi, ml-hi, or-hi and te-hi. 

\section{Model architectures}
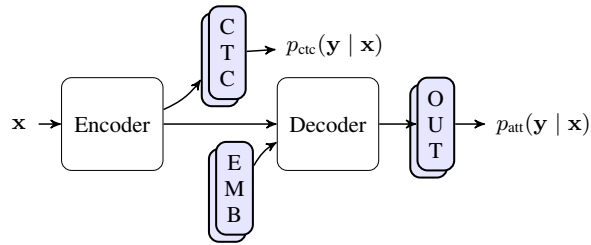
\begin{figure}[!t]
    \centering
    \resizebox{\columnwidth}{!}{%
    \begin{tikzpicture}[
        rect/.style={minimum size=1cm,text width=1.2cm,minimum height=1.3cm, fill=gray!0, align=center, 
          rectangle,draw,rounded corners},
        layer/.style={minimum height=1.3cm, text width=0.3cm, fill=blue!10, thick,align=center, rectangle,draw,rounded corners},
        light/.style={minimum height=1.3cm, text width=0.3cm, fill=blue!10, thick,align=center, rectangle,draw,rounded corners},
        var/.style={minimum size=5mm,circle},
        arr/.style={->,>=stealth',semithick},
    ]
    \node (x) [var] at (-1.3, 0) {$\mathbf{x}$};
    \node (asr) [rect] at (0, 0) {Encoder};
    \node (ctc-r) [light] at (1.5, 0.9) {};
    \node (ctc) [layer] at (1.6, 1.0) {C\\T\\C};    
    %\node (y) [var] at (3.1, 1.2) {$\mathcal{L}_{\text{ctc}}$} ;
    \node (y) [var] at (3.1, 1.1) {$p_{\text{ctc}}(\mathbf{y} \mid \mathbf{x})$} ;
    \node (emb-r) [light] at (1.6, -1) {};
    \node (emb) [layer] at (1.7, -0.9) {E\\M\\B};    
    \node (mt) [rect] at (3, 0) {Decoder};
    \node (out-r) [light] at (4.4, -0.1) {};
    \node (out) [layer] at (4.5, 0) {O\\U\\T};    
    %\node (z) [var] at (5.7, 0) {$\mathcal{L}_{\text{att}}$};
    \node (z) [var] at (6, 0) {$p_{\text{att}}(\mathbf{y} \mid \mathbf{x})$};
    \draw
    (x)   edge [arr] (asr)
    (asr) edge [arr, midway,bend right=15] (ctc-r)
    (ctc) edge [arr] (y)
    (asr) edge [arr] (mt)
    (emb) edge [arr, bend left=15] (mt)
    (mt)  edge [arr] (out)
    (out) edge [arr] (z)
    ;
    \end{tikzpicture}
    }
    \caption{Encoder-decoder architecture of the bi-lingual ASR, where $\mathbf{x}$ represents filter bank features extracted from speech signal, and $\mathbf{y}$ represents the output text sequence. Each language has a specific CTC, embedding (EMB) and output (OUT) layers. The pre-trained bi-lingual ASR is fine-tuned for ST relying on target-language-specific CTC, EMB and OUT layers.}
    \label{fig:asr_st_arch}
    \vspace{-0.3cm}
\end{figure}

This section briefly explains the model architectures of ASR, ST and baseline systems used in the study. The pre-training of ASR is always bilingual and is based on transformer joint CTC/attention architecture~\cite{joint_ctc}, except that each language has its own CTC, embedding and output layers~\cite{kesiraju_strategies_2023}. The architecture is depicted in Fig.~\ref{fig:asr_st_arch}.  The ST model shares the same architecture as the ASR model, but when initialized from the pre-trained ASR, it only uses the target language-specific CTC, embedding, and output layers. This ensures that the ST model only generates text from the target language. We employed this architecture because it yielded state-of-the-art results in IWSLT'23 Marathi$\rightarrow$Hindi low-resource ST task~\cite{iwslt:2023,kesiraju-etal-2023-systems}.

The baseline systems are based on multilingual, multimodal foundational models, SONAR~\cite{sonar:2023} and Seamless~\cite{seamless2023}. The backbone of both SONAR and Seamless is the pre-trained wav2vec-bert-2.0~\cite{w2v-bert} and the NLLB\cite{NLLB} based massive machine translation model based on NLLB \cite{NLLB}. SONAR encodes both speech and text utterances into a single vector, from which the SONAR decoder generates the hypothesis in auto-regressive fashion. 
\begin{figure*}[!ht]
    \centering
    \includegraphics[width=0.95\linewidth]{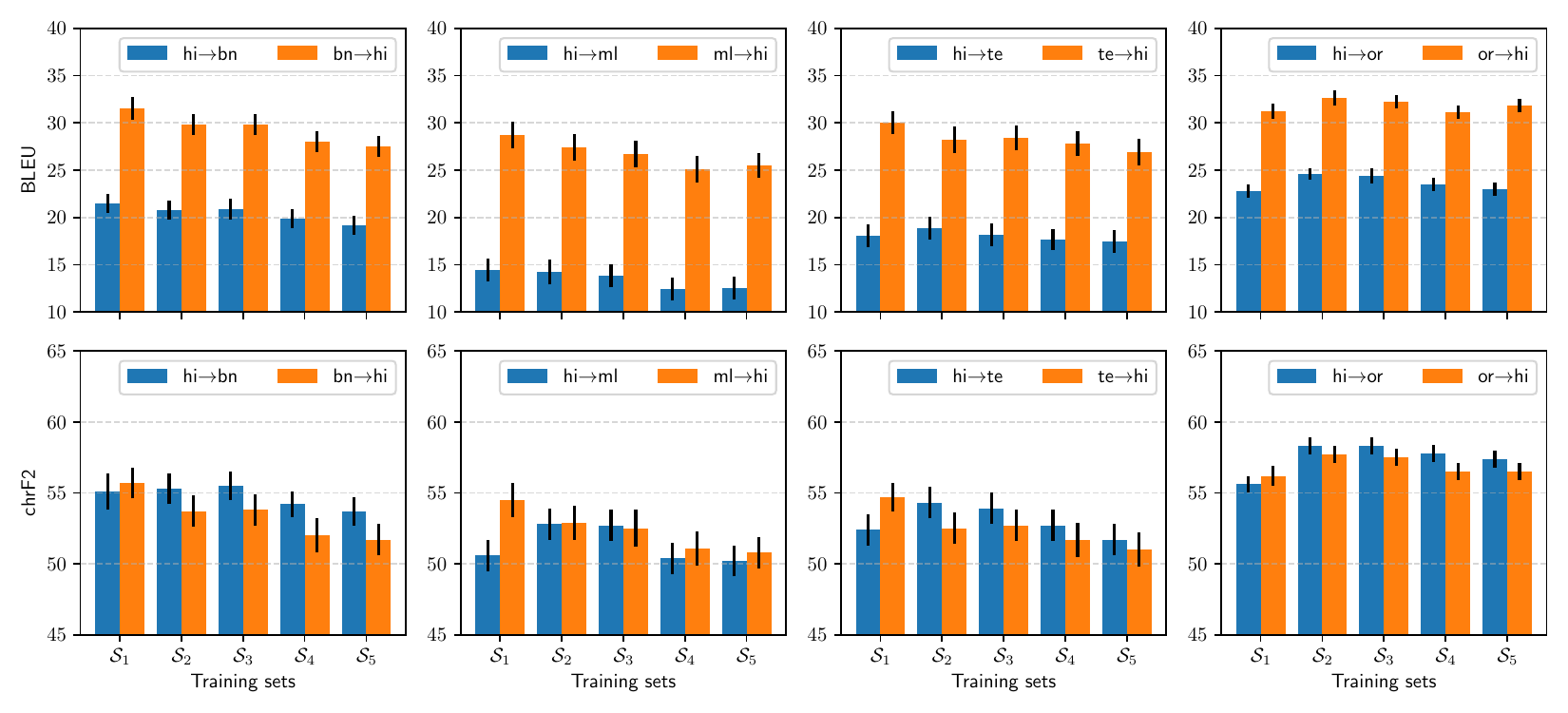}
    \caption{BLEU and ChrF2 scores along with 95\% CI for 4 language pairs, varying the quality and quantity of training data ($\mathcal{S}_1 \supset \mathcal{S}_2 \supset \mathcal{S}_3 \supset \mathcal{S}_4 \supset \mathcal{S}_5$). Table~\ref{tab:comparison_s1_s5} provides the corresponding $p$-values for the significance tests ($\mathcal{S}_1$ vs. $\mathcal{S}_5$).}
    \label{fig:slt_weak}
\end{figure*}    
    
\section{Experimental setup}
\label{sec:exp}
This section describes the details of ASR, ST and baseline models used in our experiments. The training data are augmented with speed perturbations using factors of 0.9, 1, and 1.1. From the input speech, 80-dimensional filter bank features are extracted on the fly. A sentence piece-based subword tokenizer~\cite{kudo-richardson-2018-sentencepiece} was used to build a vocabulary of 8000 tokens for each language independently. We used the ESPnet toolkit~\cite{inaguma-etal-2020-espnet} for all of our experiments, which were performed primarily on NVIDIA A100 GPUs.

\subsection{ASR \texorpdfstring{$\rightarrow$}{->} ST} 
\subsubsection{ASR pre-training}
For ASR pre-training, we used bilingual models with joint CTC/attention objective with a CTC weight of 0.3 The training was conducted over a maximum of 100 epochs with a learning rate of 0.0005, using the Adam optimizer with gradient clipping set to 5, and a warm-up scheduler with 25,000 steps. The batch size was set to 256 with folded mini-batches. Early stopping was applied with a patience of 5 epochs, and we averaged the 10 best-performing checkpoints to get the final model. On 2 GPUs, the training took an average of 30 hours. We used joint decoding using beam search with width 10. The ASR models were evaluated using word error rates (WER). We have four bilingual ASR models, bn-hi, ml-hi, or-hi and te-hi. The word error rates are presented in Table~\ref{tab:wer}. Hindi (hi) is part of each bilingual ASR system, and the reported WER is the average number. We can see from Table~\ref{tab:wer} that the Hindi-based ASR has a much lower WER compared to other ones, due to its much larger training dataset.

\begin{table}[!ht]
    \caption{WER of hi-xx bi-lingual ASR systems}
    \label{tab:wer}
    \centering
    \footnotesize
    \begin{tabular}{rrrrr} \toprule
      bn    & hi   & ml  & or  & te  \\ \midrule
      20.5 &  7.9  & 20.4 & 19.3  & 21.5 \\ \bottomrule      
    \end{tabular}
    \vspace{-0.5cm}
\end{table}

%  \begin{table}[!ht]
% \caption{Statistical significance of models trained $\mathcal{S}_1$ with $\mathcal{S}_5$. This Table should be interpreted together with Fig.~\ref{fig:slt_weak}}
% \begin{center}
% \footnotesize
% \begin{tabular}{ccc} \toprule
% % \hline
% Lang. pair& $p$-value (BLEU)& $p$-value (chrF2)\\
% \midrule
% te-hi & 0.002 & 0.002 \\
% ml-hi & 0.002 & 0.002 \\
% bn-hi & 0.002 & 0.002 \\
% or-hi & 0.011 & \textbf{0.270 }\\
% hi-te & 0.048 & 0.022 \\
% hi-ml & 0.001 & \textbf{0.089} \\
% hi-bn & 0.002 & 0.002 \\
% hi-or & \textbf{0.159} & 0.002 \\ \bottomrule
% % \hline
% \end{tabular}
% \label{tab:comparison_s1_s5}
% \end{center}
%    \vspace{-0.7cm}
% \end{table}
%\vspace{5em}
\subsubsection{Finetuning for ST}
Speech translation models are initialized with their respective bilingual ASRs and then fine-tuned independently on each of the Shrutilipi-anuvaad training splits $\{\mathcal{S}_1 \ldots \mathcal{S}_5\}$. The ST models are unidirectional. This transfer learning technique is used to enhance performance in low-resource scenarios. The ST models were optimized with joint CTC/attention objective for translation. The models were trained for a maximum of 100 epochs with a learning rate of 1e-4 and a batch size of 64. Adam optimizer was used and a CTC weight of 0.3 was applied in training and decoding. Early stopping was applied with a patience of 5 epochs. Training times for ST models varied between 2 and 6 hours on a single GPU, depending on the size of the training split. We used joint decoding using beam search with width 10.  ST models were evaluated using BLEU and chrF2 scores, along with 95\% confidence intervals, all calculated using the sacreBLEU toolkit~\cite{post-2018-call}.

\begin{table}[!ht]
\caption{$p$ values for the statistical significance test for ST models trained on $\mathcal{S}_1$ performing better than models trained on $\mathcal{S}_5$ in terms of BLEU scores. This Table should be interpreted together with Fig.~\ref{fig:slt_weak}}
\begin{center}
\scalebox{0.85}{
\begin{tabular}{rrrrrrrr} \toprule
te-hi & ml-hi & bn-hi & or-hi & hi-te & hi-ml & hi-bn & hi- or \\ \midrule 
0.002 & 0.002 & 0.002 & 0.011 & 0.048 & 0.001 & 0.002 & 0.159 \\ \bottomrule
\end{tabular}
}
\label{tab:comparison_s1_s5}
\end{center}
   \vspace{-0.7cm}
\end{table}

\subsection{Baseline models}
For our baseline, we fine-tuned Seamless (large-v2), a 2.3B parameter model. The hyperparameters used were: learning rate: 1e-5, warm-up steps: 1000, max epochs: 20, patience: 7, label smoothing: 0.2, and train batch size: 5. The baseline SONAR and Seamless models without fine-tuning are also used for machine and speech translation (MT, ST) evaluation.

\begin{table*}[!ht]
\caption{Comparison of baseline MT and ST systems with our system on the test sets. \textbf{Bold} and \underline{underlined} numbers indicate first and second highest BLEU scores. Parentheses show standard error representing confidence intervals.}
    \label{tab:slt_sonar}
    \centering
    \footnotesize
    \resizebox{\textwidth}{!}{ % Resize table to fit full width
    \begin{tabular}{clrrrrrrrr} \toprule
    System & Metric & bn$\rightarrow$hi & ml$\rightarrow$hi & or$\rightarrow$hi & te$\rightarrow$hi & hi$\rightarrow$bn & hi$\rightarrow$ml & hi$\rightarrow$or & hi$\rightarrow$te \\ \midrule  
    SONAR & BLEU & 24.8 ($\pm$0.9) & 24.0 ($\pm$1.2) & 24.6 ($\pm$0.6) & 23.7 ($\pm$1.1) & 
                    11.6 ($\pm$0.7) & 8.7  ($\pm$0.9) & 11.7 ($\pm$0.5) & 10.9 ($\pm$0.9)  \\
    MT   & chrF2 & 54.2 ($\pm$1.0) & 56.2 ($\pm$1.0) & 55.1 ($\pm$0.5) & 53.7 ($\pm$0.9) & 
                    51.3 ($\pm$1.0) & 56.1 ($\pm$0.8) & 51.6 ($\pm$0.5) & 52.8 ($\pm$0.8)\\ 
    \midrule       
     Seamless & BLEU & 20.0 ($\pm$0.8) & 19.2 ($\pm$1.1) & 22.6 ($\pm$0.6) & 19.9 ($\pm$0.9) &
                       12.1 ($\pm$0.8) & 7.1  ($\pm$0.8) & 11.6 ($\pm$0.5) & 9.9 ($\pm$0.8) \\
    MT  & chrF2  & 51.7 ($\pm$0.7) & 52.9 ($\pm$0.9) & 54.7 ($\pm$0.5) & 51.1 ($\pm$0.8) &  
                    52.3 ($\pm$0.7) & 54.4 ($\pm$0.8) & 52.1 ($\pm$0.5) & 52.0 ($\pm$0.7) \\
    \midrule 
    SONAR & BLEU & 15.7 ($\pm$0.7) & 12.1 ($\pm$0.9) & 17.6 ($\pm$0.5) & 14.8 ($\pm$0.9) &
                   7.7 ($\pm$0.5) & 5.8 ($\pm$0.7) & 8.2 ($\pm$0.4) & 7.3 ($\pm$0.8)\\
    ST  & chrF2  & 44.2 ($\pm$0.9) & 43.9 ($\pm$1.0) & 48.6 ($\pm$0.5) & 45.1 ($\pm$0.9) &
                   46.0 ($\pm$0.8) & 49.3 ($\pm$0.9) & 46.4 ($\pm$0.5) & 45.5 ($\pm$0.9)\\ 
    \midrule       
    Seamless & BLEU  & 16.6 ($\pm$0.8) & 15.5 ($\pm$1.0) & 18.5 ($\pm$0.5) & 14.5 ($\pm$0.8) &                 
                    8.1 ($\pm$0.6) & 5.6 ($\pm$0.7) & 10.0 ($\pm$0.5) & 7.7 ($\pm$0.8) \\
    ST  & chrF2  & 40.5 ($\pm$1.2) & 47.6 ($\pm$0.8) & 49.1 ($\pm$0.6) & 43.7 ($\pm$1.0) &
                   43.8 ($\pm$1.0) & 51.6 ($\pm$0.9) & 49.0 ($\pm$0.5) & 45.7 ($\pm$0.9) \\ 
    \midrule       
    Seamless & BLEU & \textbf{35.6} ($\pm$1.1) & \textbf{34.0} ($\pm$1.3) & \textbf{34.3} ($\pm$1.3) & \textbf{34.2} ($\pm$1.2) & \underline{20.0} ($\pm$1.0) & \textbf{14.7} ($\pm$1.2) & \underline{23.1} ($\pm$0.7) & \textbf{19.8} ($\pm$1.2) \\
    ST finetuned  & chrF2  & 61.0 ($\pm$0.9) & 61.5 ($\pm$0.9) & 60.7 ($\pm$0.5) & 43.7 ($\pm$1.0) & 58.3 ($\pm$1.0) & 59.0 ($\pm$0.9) & 59.8 ($\pm$0.5) & 59.1 ($\pm$0.9) \\ 
    \midrule
    Our  & BLEU   & \underline{31.5} ($\pm$1.2) & \underline{28.8} ($\pm$1.4) & \underline{32.6} ($\pm$0.8) & \underline{30.0} ($\pm$1.2) &  \textbf{21.5} ($\pm$1.0) & \underline{14.4} ($\pm$1.2) & \textbf{24.6} ($\pm$0.6) & \underline{18.6} ($\pm$1.2) \\
    ASR$\rightarrow$ST  & chrF2  & 55.7 ($\pm$1.1) & 54.5 ($\pm$1.2) & 57.7 ($\pm$0.6) & 54.7 ($\pm$1.0) & 55.1 ($\pm$1.3) & 50.6 ($\pm$1.1) & 58.3 ($\pm$0.6) & 54.1 ($\pm$1.1) \\ 
    \bottomrule
    \end{tabular}
    }
\end{table*}

\section{Results and analysis}
 \label{sec:res}

% \begin{figure*}[!ht]
%     \centering
%     \includegraphics[width=0.95\linewidth]{figures/baselines.pdf}
%     \caption{BELU and chrF2 scores of SONAR ST, SONAR MT, SM4T ST, and SM4T MT}
%     \label{fig:baselines}
% \end{figure*}
  Each of the bilingual ASRs is used to initialize a respective ST system that is trained on the splits $\{\mathcal{S}_1 \ldots \mathcal{S}_5\}$ independently. For 4 language pairs, we have 8 speech translation directions, and each ST model trained independently on 5 training splits resulted in 40 ST systems. The results of which are presented in Fig.~\ref{fig:slt_weak}. Each subplot represents ST results (BLEU scores in the first row, chrF2 in second) in both directions for a language pair. We observe that xx$\rightarrow$hi models (orange bars) generally achieve significantly higher BLEU and chrF2 scores compared to their reverse counterparts (blue bars), hi$\rightarrow$xx. These trends can be attributed to the significantly larger Hindi pre-training datasets for ASR and as a consequence a stronger internal language model for Hindi. The next trend to be observed from Fig.~\ref{fig:slt_weak} is that ST models fine-tuned on much larger but weaker sets $\mathcal{S}_1, \mathcal{S}_2$ tend to perform better than those trained on much smaller sets of better quality $\mathcal{S}_5$. We conducted statistical significance tests to find the reliability of the claim. The 95\% confidence intervals (CI) are depicted in the same Fig.~\ref{fig:slt_weak}, with corresponding $p$ values tabulated in Table~\ref{tab:comparison_s1_s5}. Another observation from Fig.~\ref{fig:slt_weak} is seen for the language pair hi-or; we see that systems trained on $\mathcal{S}_1$ perform slightly worse than systems trained on $\mathcal{S}_2$, despite having twice the amount of training data, albeit poorer in quality (similarity scores between (0.5, 0.6)). This trend is not observed for other pairs, where the difference in performance between $\mathcal{S}_1$ and $\mathcal{S}_2$ is not statistically significant. 
These experiments tell us that there exists a trade-off between the quality and quantity of weakly labeled data, where beyond a point which adding more lower quality data will cause diminishing returns (e.g. hi-or).

Next, we present the results of the speech translation systems trained from scratch using the largest training split, $\mathcal{S}_1$. These results are shown in Table~\ref{tab:slt_scratch}. It can be seen that the results are quite low, with or$\rightarrow$hi standing out, probably due to the higher training data (Table~\ref{tab:slt_data_stats}). Comparison of Table~\ref{tab:slt_scratch} and Fig.~\ref{fig:slt_weak} shows the benefits of (in-domain) pre-training.

\begin{table}[!ht]
\caption{Results of ST systems trained from scratch on $\mathcal{S}_1$ split.}
    \label{tab:slt_scratch}
    \centering
    \resizebox{80mm}{!}{ % Scale to fit within column width
    \begin{tabular}{c p{1.5cm} p{1.5cm} p{1.5cm} p{1.5cm}} \toprule
    & bn$\rightarrow$hi & ml$\rightarrow$hi & or$\rightarrow$hi & te$\rightarrow$hi \\ \midrule
     BLEU    & 6.8 ($\pm$0.7) & 9.3 ($\pm$0.8) &  14.7 ($\pm$0.5) & 11.8 ($\pm$0.9) \\
     chrF2   & 22.7 ($\pm$0.8)  & 29.4 ($\pm$1.1) & 35.3 ($\pm$0.6) & 31.5 ($\pm$1.0) \\ \bottomrule
    \end{tabular}
    }
\end{table}

% Here, we present the performance of  end-to-end spoken language translation systems (SLTs) on different versions of data, comparing them to analyse quality vs quantity tradeoffs. Given that the Bleu and Chrf2 scores are close, we have calculated confidence intervals, and statistical significance values for each of the versions. Statistical significance tests were conducted using the bootstrapping method~\cite{koehn-2004-statistical} with 500 samples.

% Discuss the scores and analyse

Table~\ref{tab:slt_sonar} compares our best ST system with SONAR and Seamless MT and ST, and Seamless fine-tuned ST systems across all language pairs. We can see from the first two rows that the SONAR MT systems perform better than the Seamless MT systems, whereas the Seamless ST system outperformed the SONAR ST systems (rows 3-4). Although these systems exhibit lower BLEU scores, their chrF2 scores perform better. This suggests that baseline models are able to correctly translate several words or subwords but struggle to translate higher-order $n$-grams in the right order, which explains the lower BLEU scores. The last two rows from Table~\ref{tab:slt_sonar} compares the ST system fine-tuned on the curated corpus. Seamless ST fine-tuned (row 5) achieves superior BLEU scores in 6 out of 8 directions, while our ST model (row 6) achieves the best BLEU scores in 2 out of 8 directions. We observed that the Seamless model gave the best results when fine-tuned on the smaller but high-quality $\mathcal{S}_5$ set, whereas our model gave best results when fine-tuned on the bigger training split $\mathcal{S}_1$. 
%We hypothesize that since Seamless is already a speech translation model, training on lower amounts of high quality data resulted in better scores, and adding more data of lower quality degraded this performance. Whereas our models are initialized from ASR systems, require more training data to learn to translate, hence a larger amount of weakly labeled data is beneficial.
% Across language pairs, our ASR$\rightarrow$ST model fine-tuned on large but weakly labeled data $\mathcal{S}_1$ achieves BLEU scores 3-6 points lower than the fine-tuned Seamless model - showing results after training on a smaller but high-quality dataset $\mathcal{S}_5$. 
The performance gap is smallest for hi-bn (21.5 vs. 20.0), where our model outperforms Seamless, and largest for ml-hi (28.8 vs. 34.0), indicating that this pair benefits more from high-quality data. 
%In terms of chrF2, the fine-tuned baseline consistently surpasses our model by 4-6 points, with the largest differences observed in ml-hi (54.5 vs. 61.5 chrF2) and bn-hi (55.7 vs. 61.0 chrF2). 
For Hindi-to-regional pairs (hi-bn, hi-or, hi-te, hi-ml), where translation is generally more challenging, our model remains competitive despite being trained under weaker supervision, achieving BLEU scores in a similar range compared to Seamless (14.4 vs 14.7 and 24.6 vs 23.1).
These results suggest that, while fine-tuning on high-quality data leads to superior performance, large-scale weak supervision remains a viable alternative for ST in low-resource settings.

% The SLT versions are also compared to the text-to-text and Speech-to-text baselines generated using SONAR and SEAMLESS. Additionally, ASR initialized SLTs are also compared to from scratch SLTs. The results show that from scratch SLTs perform significantly worse than ASR initialized SLTs in low0resource scenarios. 

% Table - SLT results from scratch training.

\section{Conclusion and future work}
\label{sec:concl}
In this paper, we aimed to train end-to-end speech translation (ST) systems by using automatically curated weakly labeled data. To this extent, we built our ST corpus, \textit{Shrutilipi-anuvaad}, based on the multilingual Shrutilipi corpus by automatically mining parallel texts using the SONAR sentence encoder. We have conducted extensive experiments to analyze the effect of the quality and quantity of weakly labeled data on the performance of ST systems. We also observed significant improvements by employing transfer learning, i.e., initializing ST from pre-trained bilingual ASR systems. We also benchmarked and compared state-of-the-art multilingual speech translation models on the created test sets. Finally, we release the data splits for the curated corpus, there by contributing free and potentially useful datasets to the research community.

In the near future, we aim to extend the Shrutilipi-anuvaad dataset with additional languages and build ST systems for additional Indic-to-Indic directions.

\section{Limitations}
All our datasets are sourced from single-domain radio broadcasts resulting in a consistent speaking style across training and evaluation. Additionally, our pre-trained models are also derived from the same domain, which may influence performance. 

\section{Acknowledgments}
Santosh Kesiraju was supported by the Ministry of Education, Youth and Sports of the Czech Republic (MoE) through the OP JAK project ``Linguistics, Artificial Intelligence and Language and Speech Technologies: from Research to Applications'' (ID:CZ.02.01.01/00/23\_020/0008518). We also acknowledge Kakinada Institute of Engineering and Technology (KIET), for providing the computational resources used in this work.

% \newpage
\begingroup
\footnotesize
\bibliographystyle{ieeetr}
\bibliography{refs}
\endgroup

%\section{References}

\end{document}